\documentclass[electronic]{vgtc}                 

\ifpdf
  \pdfoutput=1\relax                   
  \usepackage{graphicx}                
  \DeclareGraphicsExtensions{.pdf,.png,.jpg,.jpeg} 
\else
  \usepackage{graphicx}                
  \DeclareGraphicsExtensions{.eps}     
\fi%

\graphicspath{{figures/}{pictures/}{images/}{./}} 

\usepackage{microtype}                 
\PassOptionsToPackage{warn}{textcomp}  
\usepackage{textcomp}                  
\usepackage{mathptmx}                  
\usepackage{amsmath}
\usepackage{amssymb}
\usepackage{float}
\usepackage{multirow}

\usepackage[ruled,vlined]{algorithm2e}
\usepackage{times}                     
\usepackage{cite}                      
\usepackage{tabu}                      
\usepackage{booktabs}                  

\onlineid{1352}

\vgtccategory{Research}

\vgtcinsertpkg

\title{GazeMoDiff: Gaze-guided Diffusion Model for\\ Stochastic Human Motion Prediction}

\author{Haodong Yan\thanks{e-mail: k610215095@stu.xjtu.edu.cn}\\ %
        \scriptsize Xi'an Jiaotong University %
\and Zhiming Hu\thanks{e-mail: zhiming.hu@vis.uni-stuttgart.de}\\ %
        \scriptsize University of Stuttgart %
\and Syn Schmitt\thanks{e-mail: schmitt@simtech.uni-stuttgart.de}\\ %
     \scriptsize University of Stuttgart %
\and Andreas Bulling\thanks{e-mail: andreas.bulling@vis.uni-stuttgart.de}\\ %
     \scriptsize University of Stuttgart}

\teaser{
 \centering
  \includegraphics[width=\linewidth]{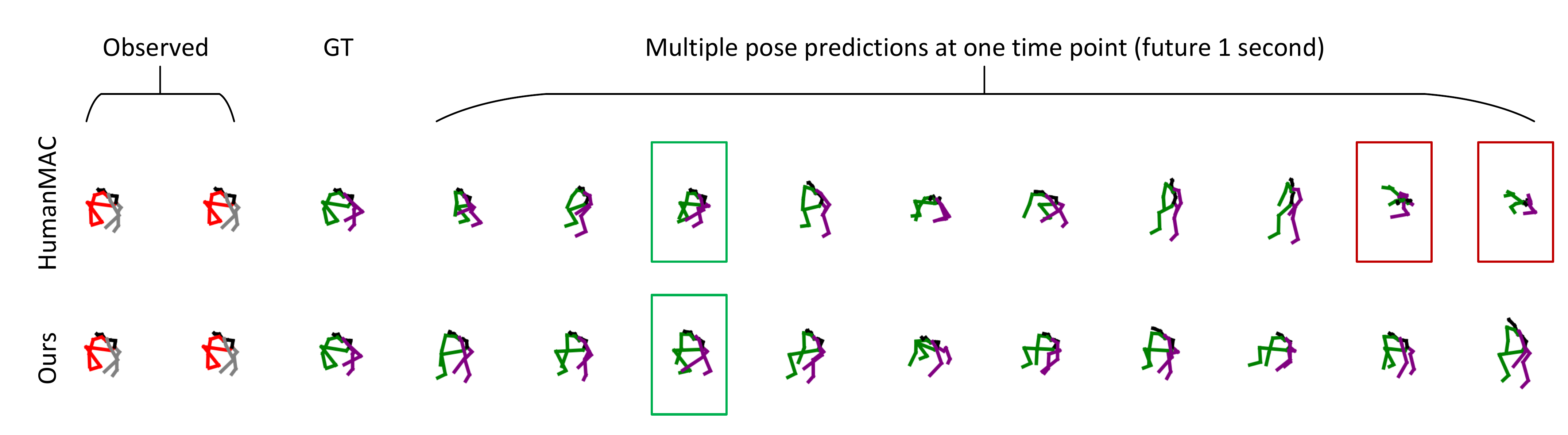}
  \caption{Ground truth (GT) human pose at future one second and multiple pose predictions generated by different methods 
  on the GIMO dataset~\cite{zheng2022gimo} with the best prediction (lowest $l_2$ distance to GT) boxed in green and implausible cases marked in red. Our method can generate multiple reasonable motions while the state-of-the-art method HumanMAC~\cite{chen2023humanmac} produces some implausible predictions.
  }
  \label{fig:teaser}   
  
}

\def\shownotes{1}

\ifnum\shownotes=1
\newcommand\zhiming[1]{\textcolor{cyan}{Zhiming: #1}}

\newcommand\todo[1]{\textcolor{red}{#1}}
\newcommand\reply[1]{\textcolor{blue}{Relpy: #1}}
\newcommand\andreas[1]{\textcolor{orange}{Andreas: #1}}
\else 
\newcommand\andreas[1]{}
\newcommand\zhiming[1]{}

\newcommand\todo[1]{}
\newcommand\reply[1]{}
\fi

\abstract{
Human motion prediction is important for many virtual and augmented reality (VR/AR)
applications such as collision avoidance and realistic avatar generation.
Existing methods have synthesised body motion only from observed past motion, despite the fact that human eye gaze is known to correlate strongly with body movements and is readily available in recent VR/AR headsets.
We present \textit{GazeMoDiff} -- a novel gaze-guided denoising diffusion model to generate stochastic human motions.
Our method first uses a gaze encoder and a motion encoder to extract the gaze and motion features respectively, then employs a graph attention network to fuse these features, and finally injects the gaze-motion features into a noise prediction network via a cross-attention mechanism to progressively generate multiple reasonable human motions in the future.
Extensive experiments on the MoGaze and GIMO datasets demonstrate that our method outperforms the state-of-the-art methods by a large margin in terms of 
multi-modal final displacement error (17.3\% on MoGaze and 13.3\% on GIMO).
We further conducted a human study (N=21) and validated that the motions generated by our method were perceived as both more precise and more realistic than those of prior methods.
Taken together, these results reveal the significant information content available in eye gaze for stochastic human motion prediction as well as the effectiveness of our method in exploiting this information.
}

\CCScatlist{
  \CCScatTwelve{Human-centred computing}{Human computer interaction}{Interaction paradigms}{Virtual reality};
  \CCScatTwelve{Computing methodologies}{Machine learning}{Machine learning approaches}{Neural networks};
}

\begin{document}



\firstsection{Introduction}
\maketitle

Generating realistic human body movements is a key research challenge in the area of virtual and augmented reality (VR/AR) and is the basis for safe, smooth, and immersive human-environment~\cite{david2021towards, sun2018towards} and human-human interactions~\cite{kim2017activity,holm2012collision}.
Human motion prediction (HMP) enables a number of exciting applications such as redirected walking to create the illusion of unlimited virtual interaction spaces~\cite{azmandian2016automated,sun2018towards} or to steer users away from physical boundaries, such as walls, and thus avoid collisions~\cite{gamage2021so,zheng2022gimo}.
Human motion prediction can also provide users with a low-latency experience by preparing VR content in advance based on the predicted future human poses~\cite{hou2019head,9983833} and it has been used to generate human-like motions for virtual agents to enhance the interaction experience~\cite{bhattacharya2021text2gestures,du2023avatars}.

Previous work on HMP has typically generated human motions in a deterministic way, i.e. by producing only a single prediction at a time~\cite{martinez2017human,butepage2017deep,cao2020long,gopalakrishnan2019neural}.
Recently, in light of the fact that human motion is stochastic by nature~\cite{chen2023humanmac, barquero2022belfusion}, researchers have turned to stochastic human motion prediction, i.e. generating a number of reasonable human motions at a time~\cite{chen2023humanmac,yuan2020dlow,barquero2022belfusion,barsoum2018hp}.
Stochastic human motion prediction suits the needs of many VR applications.
For example, to minimise the collision risk in a virtual environment, it is necessary to consider multiple possible future trajectories to warn the users.
To produce realistic virtual agents, it is also beneficial to synthesise multiple reasonable human motions that users can peruse and select from according to their personal preferences.
However, existing stochastic HMP methods typically generate human motions using only past observed motions and neglect other modalities, in particular human eye gaze.
With rapid advances in eye tracking technology, human eye gaze information has become readily available in many VR/AR head mounted displays (HMDs), such as HTC Vive Pro Eye, Varjo VR-3, and Vision Pro, and has demonstrated its potential for gaze-based interaction~\cite{sidenmark2019selection} and gaze-contingent rendering~\cite{hu2019sgaze, hu2020dgaze} in VR/AR.
In addition, a large body of work in the cognitive sciences and human-centred computing has shown that human body movements are closely coordinated with human gaze behaviour~\cite{hu2021ehtask,hu2021fixationnet, sidenmark2019eye, freedman2008coordination}.
Despite this close coordination, information on eye gaze has not been used for stochastic human motion prediction so far.

To address this limitation we propose \textit{GazeMoDiff} -- the first \textbf{Gaze}-guided human \textbf{Mo}tion \textbf{Diff}usion model to generate multiple reasonable human motions.
Our method first uses a 1D convolutional neural network (1D CNN) and a graph attention network (GAT) to extract features from eye gaze and body motion respectively, then employs a spatio-temporal graph attention network to fuse these features, and finally injects the gaze-motion features into a noise prediction network via a cross-attention mechanism to generate multiple reasonable human future motions through a progressive denoising process.
We extensively evaluate our method on the MoGaze dataset~\cite{kratzer2020mogaze} for real-world setting as well as on the GIMO dataset~\cite{zheng2022gimo} for AR setting.
Experimental results demonstrate that our method significantly outperforms the state-of-the-art methods that only use past body poses, achieving an improvement of 
16.7\% on MoGaze and 10.8\% on GIMO in multi-modal average displacement error and 17.3\% on MoGaze and 13.3\% on GIMO in multi-modal final displacement error.
To qualitatively evaluate our method, we further conduct a user study and the responses from $23$ users validate that the motions generated by our method are perceived as both more precise and more realistic than predictions of prior methods.
The full source code and trained models are available at zhiminghu.net/yan24\_gazemodiff.

The main contributions of our work are three-fold:
\begin{itemize}
    \item We propose a novel gaze-guided diffusion model for stochastic human motion prediction that uses a spatio-temporal graph attention network to fuse the gaze and motion features and then injects these features into a noise prediction network via a cross-attention mechanism to generate multiple reasonable human motions in the future.
    
    \item We conduct extensive experiments on two public datasets for both real-world and AR settings and demonstrate significant performance improvements over state of the art.
    
    \item We report a user study that shows the motions generated by our method are perceived as more precise and more realistic than that from prior methods.
\end{itemize}
\section{Related Work}

\subsection{Human Motion Prediction}
Human motion prediction is a fundamental research topic in virtual and augmented reality.
Early studies commonly considered HMP as a deterministic sequence prediction task, tackling it with recurrent neural network (RNN)~\cite{martinez2017human,ghosh2017learning,li2017auto,fragkiadaki2015recurrent},
graph neural network (GNN)~\cite{li2021directed,li2021multiscale, hu24hoimotion} and Transformers~\cite{aksan2021spatio,martinez2021pose}. 
Noticing that human body motions are inherently stochastic, recent works began to predict human motions in a stochastic way using generative models such as variational autoencoder (VAE)~\cite{aliakbarian2021contextually,mao2021generating}, generative adversarial network (GAN)~\cite{kundu2019bihmp,barsoum2018hp,jain2020gan}, and Flow networks~\cite{yuan2020dlow}.
These methods can generate diverse human motions through a diversity-aware loss or sampling strategy~\cite{mao2021generating,yuan2020dlow}, but their predictions are not physically plausible.
Inspired by the recent success of denoising diffusion models in the field of image generation~\cite{ho2020denoising,ramesh2022hierarchical,nichol2021glide,zhang2023adding}, researchers have adopted diffusion models to the task of stochastic human motion prediction and have achieved more realistic predictions than traditional generative approaches~\cite{barquero2022belfusion,chen2023humanmac}.
However, existing methods only generate future motion predictions based on observed past motion, neglecting the fact that human eye gaze is known to correlate strongly with body movements and is readily available in recent VR/AR headsets.
To fill this gap, in this work we propose the first gaze-guided diffusion model for stochastic human motion prediction.

\subsection{Correlations between Eye Gaze and Body Motion}

Intuitively, a period of human eye gaze information seems related to human intention~\cite{koochaki2018predicting}, which can drive future motion trends. Extensive works in cognitive science and human-centred computing have also demonstrated this strong correlation between human eye gaze and subsequent body motion. 
Some researchers have revealed that in many everyday activities, such as free viewing or object searching, human head movements are closely associated with eye movements~\cite{zangemeister1982gaze,hu2019sgaze,kothari2020gaze, hu24pose2gaze}.
Emery et al. further explored the coordination of human eye, hand, and head movements in virtual environments and leveraged this coordination to improve the performance of gaze estimation~\cite{emery2021openneeds}.
Sidenmark et al. identified the coordinated movements between eye, head, and torso during gaze shifts in virtual reality~\cite{sidenmark2020bimodalgaze,sidenmark2019eye}.
Despite the strong link between human eye gaze and body motion, prior work has only explored to use eye gaze to filter information from the full 3D scene environment and then employed the scene features for deterministic human motion prediction~\cite{zheng2022gimo}.
We are the first to directly leverage eye gaze information (without requiring information about the full 3D environment) to further boost the performance of stochastic human motion prediction.



\subsection{Denoising Diffusion Models}

\begin{figure}[tb]
 \centering 
 \includegraphics[width=\columnwidth]{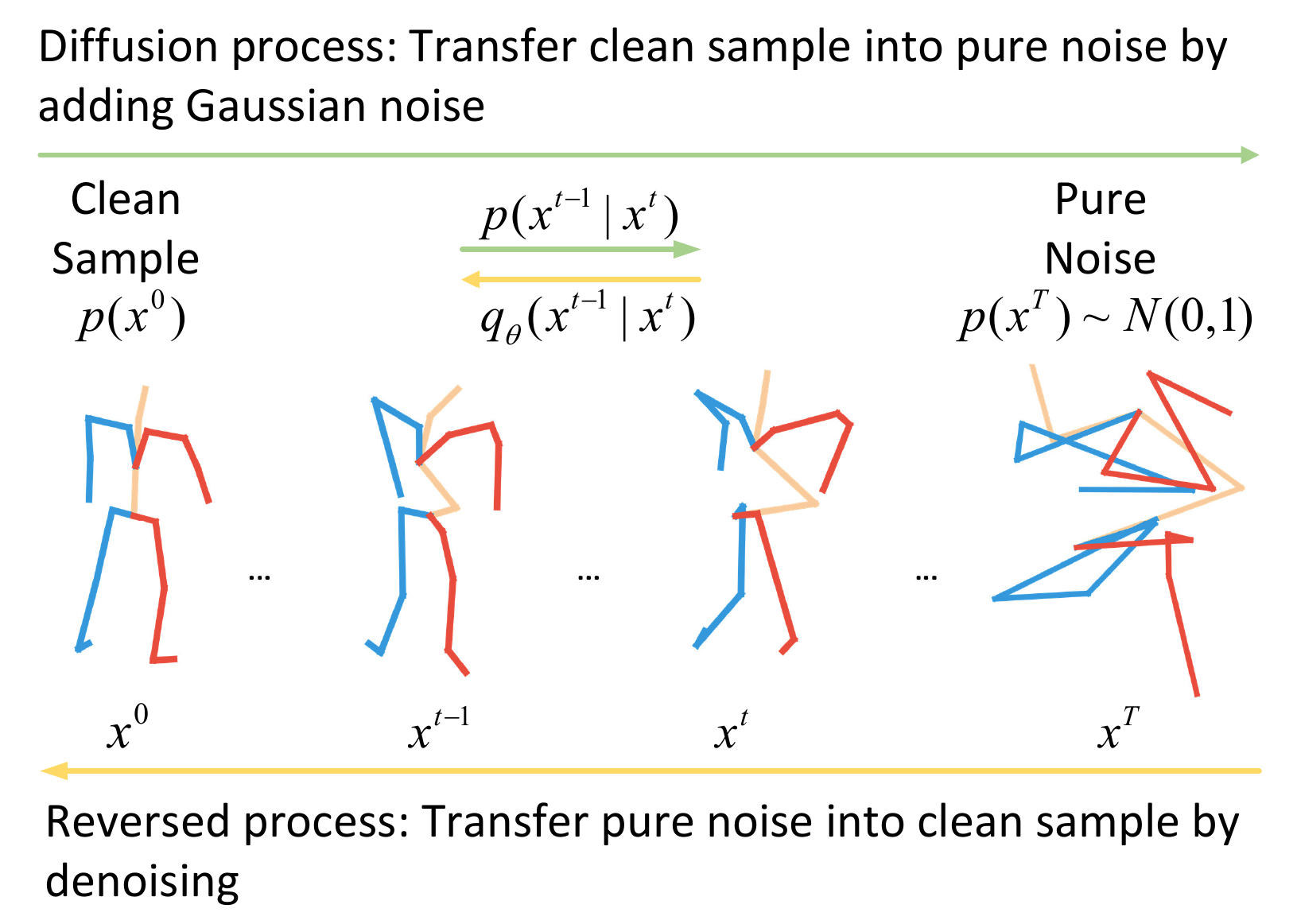}
 \caption{Diffusion process and reversed process in DDPM. 
 }
 \label{illustration_diffusion}
\end{figure}

Denoising diffusion models, or more precisely, denoising diffusion probabilistic models (DDPM)~\cite{ho2020denoising,song2020denoising,dhariwal2021diffusion,rombach2022high} are a group of the most ingenious generative models.
They aim to model reversing a Markov chain of the diffusion process illustrated in~\autoref{illustration_diffusion}. 
During training, noisy samples are obtained by incrementally adding noise to raw samples. The DDPM model then progressively reverses the diffusion process by predicting noise and denoising the samples. 
The loss is computed as the difference between the predicted noise and the Gaussian noise added during the diffusion process. 
In the inference stage, given a well-trained DDPM, it can generate realistic motions from a Gaussian distribution via the reversed diffusion process.

Owing to their power of generating realistic and high-quality samples, diffusion models have been applied in image/video generation\cite{rombach2022high,ho2022video}, anomaly detection\cite{wyatt2022anoddpm}, objection detection\cite{chen2022diffusiondet}, 3D reconstruction\cite{xu2022dream3d}, time series forecasting\cite{rasul2021autoregressive} and imputation\cite{tashiro2021csdi}. 
In a similar context, Tevet et al.\cite{tevet2022human} proposed a text-driven human motion synthesis method with diffusion models. 
Recently, Barquero et al. \cite{barquero2022belfusion} utilised a latent diffusion model to sample the diverse behaviour code to predict stochastic motions. 
Chen et al.\cite{chen2023humanmac} presented an end-to-end motion prediction framework based on diffusion models without complicated loss constraints and training processes. 
However, existing methods usually produce unrealistic predictions, probably because they only focus on past human poses and neglect important information from other modalities.
To address this limitation, our method injects gaze-motion features into the denoising step through a cross-modal attention mechanism to generate more reasonable motion predictions.


\section{GazeMoDiff Model}

\begin{figure*}[tb]
 \centering 
 \includegraphics[width=2\columnwidth]{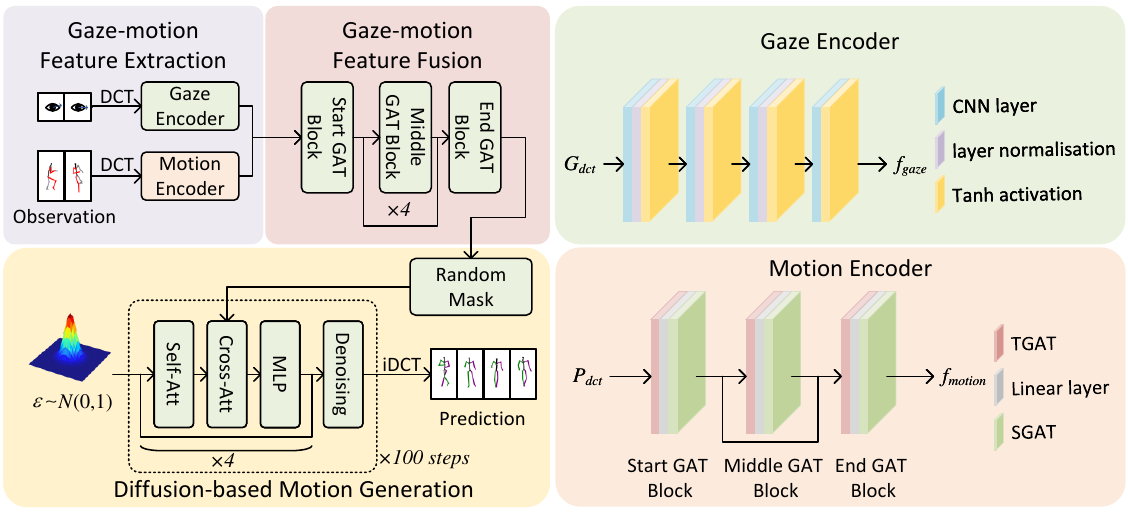}
 \caption{Overview of the proposed method \textit{GazeMoDiff}.
 \textit{GazeMoDiff} first uses a gaze encoder and a motion encoder to extract the gaze and motion features respectively, then employs a spatio-temporal graph attention network to fuse these features, and finally injects the gaze-motion features into a noise prediction network via a cross-attention mechanism to generate multiple reasonable human future motions through a progressive denoising process.
 }
 \label{fig:pipeline}
\end{figure*}

Our diffusion model for gaze-guided stochastic human motion prediction takes past body poses and eye gaze information as input.
We represent human pose $p \in \mathbb{R}^{j \times 3} $ using the 3D positions of all human joints, where $j$ denotes the number of human joints.
Eye gaze $g \in \mathbb{R}^{1 \times 3}$ is defined as a unit direction vector. 
Given $H$ frames of observed sequence $\mathbf{x} = [ (g_1, p_1 ), (g_2, p_2),...,(g_H, p_H) ]$, our goal is to predict human motions in the future $F$ frames.
Considering the stochastic nature of human motion, we generate $k$ different future motion trajectories to provide multiple reasonable predictions $\mathbf{P} = \{ \mathbf{p}_{0}, \mathbf{p}_{1},..., \mathbf{p}_{k} \}$, where $\mathbf{p}_{i} = [p_{H+1}, p_{H+2}, ..., p_{H+F}]$. 
An overview of our model is shown in \autoref{fig:pipeline}.
Our model consists of three modules: a gaze-motion feature extraction module that uses a gaze encoder and a motion encoder to extract gaze and motion features respectively, a gaze-motion feature fusion module that fuses the gaze-motion features using a spatio-temporal graph attention network, as well as a diffusion-based motion generation module that employs a cross-attention mechanism to inject the gaze-motion features into a noise prediction network to generate multiple reasonable human future motions through progressive denoising.

\subsection{Gaze-motion Feature Extraction}

\subsubsection{Gaze Encoder}
We first pad raw observed gaze data $G_{raw} \in\mathbb{R}^{{3}\times{1} \times {H}}$ to full-length gaze data $G \in\mathbb{R}^{{3}\times{1} \times {(H+F)}}$ following prior work ~\cite{chen2023humanmac}.
We then employed an approximate discrete cosine transform (DCT)~\cite{mao2021generating} that selected the first $L$ components to process the full-length gaze data $G \in\mathbb{R}^{{3}\times{1} \times {(H+F)}}$.
By leveraging this transformation, the time dimension of eye gaze signals was reduced from $H+F$ to $L$, which both improved the smoothness of generation and reduced computational complexity~\cite{mao2019learning}. 
We further employed a 1D convolutional neural network as an encoder to extract features from the transformed eye gaze sequence $G_{dct} \in\mathbb{R}^{{3}\times{1} \times {L}}$.
Specifically, we used four CNN layers with the kernel size of three.
The channel size of the first three layers is $32$ and each layer is followed by a layer normalisation (LN) and a Tanh activation function.
After the three CNN layers, we employed an extra CNN layer with three channels and a Tanh activation function to obtain the final eye gaze feature $f_{gaze}\in\mathbb{R}^{{3}\times{1} \times {L}}$.

\subsubsection{Motion Encoder}
Recent works have demonstrated that a graph structure is particularly effective for learning 
inherent correlations between human body joints~\cite{li2021directed,li2021multiscale,ma2022progressively}.
Inspired by these works, 
we treated human motions as a fully-connected spatio-temporal graph $P \in \mathbb{R}^{3 \times j \times (H+F)}$, where later $F$ frames of poses were padded with the last observed frame. 
Similarly, we employed DCT to obtain the transformed spatio-temporal graph $P_{dct} \in\mathbb{R}^{3 \times j \times L}$ for smoother generation and lower computational complexity. 
We designed a GAT-based motion encoder to model the spatio-temporal correlations in the motion sequence for extracting motion features $f_{motion}\in\mathbb{R}^{{3} \times j \times {L}}$.
The motion encoder consisted of three GAT blocks i.e. a
start GAT block, a middle GAT block, and an end GAT block. 

\textbf{Start GAT Block}. 
Given the transformed spatio-temporal graph $P_{dct}\in\mathbb{R}^{3 \times j \times L}$,  
we first enhance the features using a multi-head self-attention mechanism via the temporal GAT layer
and obtained $H^{\prime} =[{h}_1^{\prime},{h}_2^{\prime},...,{h}_L^{\prime}] \in\mathbb{R}^{3 \times j \times L}$ in the following way:

\begin{equation}
    {h}_i^{\prime}=\operatorname{LeakyReLU}\left(\frac{1}{N_{head}} \sum_{n=1}^{N_{head}} \sum_{k =1}^{L} \alpha_{i k}^n  {h}_k\right),
\end{equation}
where $ {h}_i^{\prime} \in\mathbb{R}^{3 \times j} $ is the output feature of node $i$, ${h}_k$ is the input feature of node $k$, and $N_{head}=8$ denotes the number of heads for attention.
We fused different output features from each head by averaging them.
For each head, the attention matrix $\alpha_{i k}^n$ represents interactions between each timestamp, calculated as follows: 
\begin{equation}
    \alpha_{i k}^n =\frac{\exp \left(\operatorname{LeakyReLU}\left({\mathbf{a}^n}\left[ {h}_i \oplus  {h}_k\right]\right)\right)}{\sum_{l=1}^{T} \exp \left(\operatorname{LeakyReLU}\left({\mathbf{a}^n}\left[ {h}_i \oplus  {h}_l\right]\right)\right)},
\end{equation}
where ${\mathbf{a}^n}$ is a parameter vector $\in \mathbb{R}^{6 \cdot j \times 1}$ and $\oplus$ denotes concatenation operation. 

A linear layer was then used to transfer feature dimension from $H^{\prime} \in\mathbb{R}^{3 \times j \times L}$ to $\Bar{H}^{\prime} \in\mathbb{R}^{16 \times j \times L}$.
We further proposed a spatial graph attention network layer to extract features between different joints and obtain the output of the start GAT block $H_{sta} \in\mathbb{R}^{16 \times j \times L}$. Specifically, the spatial GAT layer was similar to the temporal GAT layer
, differing primarily in how attention coefficients were computed. Rather than extracting features in the temporal dimension as in the temporal GAT, the spatial GAT extracts features across the spatial dimension. 

\textbf{Middle GAT Block}.
We designed a middle GAT block to further extract the body pose features. Taking the output of the Start GAT block $H_{sta} \in\mathbb{R}^{16 \times j \times L}$ as input, we first duplicated the temporal features for enhancement~\cite{ma2022progressively} $\left(\mathbb{R}^{16 \times j \times L}\rightarrow \mathbb{R}^{16 \times j \times 2L} \right)$. Then we used a temporal GAT layer, a linear layer, and a spatial GAT layer to further learn correlations between joints and extract features. Note that the input and output dimensions of the linear layer in the middle GAT block are both 16. Then a layer normalisation, a Tanh
activation function, and a dropout layer with 0.3 dropout rate was used after the spatial GAT layer to avoid overfitting. A skip connection was added to improve the network flow.
To keep the temporal dimension of the original input to this block the same, the middle GAT block's output $f_{mid}$ was divided in half in the temporal dimension $\left(\mathbb{R}^{16 \times j \times 2L}\rightarrow \mathbb{R}^{16 \times j \times L} \right)$.

\textbf{End GAT Block}. The end GAT block was used to obtain the final motion features with the same feature dimension in gaze features $f_{gaze}$ for further fusion. This block consists of a temporal GAT, a linear layer (reducing feature dimension from 16 to 3), a spatial GAT, a layer normalisation (LN), a Tanh
activation function, and a dropout layer with 0.3 dropout rate.
Given $f_{mid} \in \mathbb{R}^{16 \times j \times L}$, the end GAT block outputs $f_{motion}\in \mathbb{R}^{3 \times j \times L}$.

\subsection{Gaze-motion Feature Fusion}


To fuse gaze and motion features, we treated gaze and motion features as ``virtual joints'' and combined gaze and poses together as a fully connected graph $f_{in} \in \mathbb{R}^{3 \times (j+1) \times L}$. Then a similar spatio-temporal GAT was employed to fuse gaze and motion features.

The gaze-motion fusion module was composed of a start GAT block, four middle GAT blocks, and an end GAT block. The structures of those blocks are the same as the blocks in the motion encoder.
The gaze-motion features $f_{fus} \in \mathbb{R}^{3 \times (j+1) \times L}$ were calculated as:
\begin{equation}
    \label{gazemotionfuse}
    f_{fus} = GazePoseFuse \left(f_{in}\right).
\end{equation}

\subsection{Diffusion-based Motion Generation}
We formulated the motion generation task as an inverse diffusion process, which contains iterative noise prediction and denoising steps.

\textbf{Noise prediction.} Inspired by text-driven noise prediction network~\cite{zhang2022motiondiffuse}, we designed a noise prediction network
${\boldsymbol{\epsilon}_\theta}$ that employed the cross attention mechanism to guide the denoising process using the gaze-motion features. This network is developed to predict the noise at the current time step given the gaze-motion features $f_{fus}$ and noisy motion sequence $Y^t\in\mathbb{R}^{3 \times (j+1) \times L}$ from the previous time step. As illustrated in~\autoref{fig:pipeline}, ${\boldsymbol{\epsilon}_\theta}$ consists of $n$ stacked self-attention block, cross-attention block and multi-layer perception (MLP) block with skip connections.
 Additionally, to trade off diversity and realism of predictions, we train our model using classifier-free guidance~\cite{ho2022classifier}. 
Specifically, the model predicted noises from both conditional and unconditional prior by randomly setting $c = None$ for 10\% possibility and $c = f_{fus}$ for 90\% possibility. 
The predicted noise at time step $t$ is formulated as follows:
\begin{equation}
   \label{nosieprediciton}
   \Bar{\boldsymbol{\epsilon}} =  \boldsymbol{\epsilon}_{\boldsymbol{\theta}}\left( Y^t, c, t\right).
\end{equation}

To better model the temporal correlations within the noisy gaze-motion sequence $Y^t \in \mathbb{R}^{3 \times (j+1) \times L}$, we first employed a linear transformation to project the sequence into a higher-dimension latent space $Y^{\prime t} \in \mathbb{R}^{3 \times (j+1) \times L^{\prime}}$. We then applied an efficient self-attention block\cite{zhang2022motiondiffuse} to further model temporal correlations between each frame, $\mathbf{Y} = \operatorname{self-att}\left( Y^{\prime t} \right) +  Y^{\prime t}$

We then applied a step hint module to inform about how many steps of noise have been added thus far. We first obtained the timestep embedding $\mathbf{e}_t$ using position embedding\cite{vaswani2017attention}. The gaze-motion historical features $c$ were also fused with a learned linear projection of the timestep embedding, $\mathbf{e} = \mathbf{e}_t + \mathbf{W}^\prime c$. This fused embedding $\mathbf{e}$ was then injected into the output of the self-attention block, $\mathbf{Y}^{\prime} = \operatorname{step-hint}\left(\mathbf{Y},\mathbf{e}\right)$.


To incorporate the historical gaze-motion features
$f_{fus}$
and leverage their impact on noise prediction, we utilised cross-attention blocks. These blocks enable a deeper exploration of how the historical features influence the noise predictions at different denoising steps. In addition, computing the contributions of different attention heads in parallel better integrates information from both modalities (gaze and motion). 
Potentially, the noisy motions can also provide feedback to update $c$, leading to a collaborative learning process. In the cross-attention block, $\mathbf{Q}_c$  and $\mathbf{V}_c$ were calculated by gaze-motion historical features $c$ while $\mathbf{K}_c$ was calculated using the output of self-attention blocks $\mathbf{Y}^\prime$. The output of the cross-attention block was calculated as follows:
\begin{equation}
  \mathbf{Y}_c = \operatorname{Dropout}\left(\operatorname{softmax}\left( \mathbf{Q}_c \right) \operatorname{softmax}\left( \mathbf{K}_c^\top \right)\right)\operatorname{LN}\left(\mathbf{V}_c\right)+ \mathbf{Y}^\prime.
\end{equation}
\begin{equation}
     \mathbf{Q}_c = \mathbf{W}^\prime_q  c, \mathbf{K}_c = \mathbf{W}^\prime_k  \mathbf{Y}^\prime,\mathbf{V}_c = \mathbf{W}^\prime_v  c.
\end{equation}

Finally, we employed a two-layer MLP with GELU activation function and a dropout layer with 0.2 dropout rate 
to further extract features. A linear layer  
was added at the end to align with the noise dimension.


In our implementation, the noise prediction network contained four self-attention blocks, four cross-attention blocks, and four MLP blocks. Each attention operation used eight attention heads. The timestep embedding dimension and the latent space dimension were both set to 512.

\textbf{Denoising.} 
In a vanilla approach, the denoised sequence was generated directly from the Gaussian noise input. 
However, due to the accumulation of prediction noise errors, the observation information in the padded sequence is far from the truth in the latter steps. 
In each denoising step, the observed sequence was also available to guide the generation in the original space. Thus, we employed an ingenious prediction mask mechanism~\cite{chen2023humanmac} to obtain denoised motions $\mathbf{\Bar{p}} \in \mathbb{R}^{3 \times j \times F}$ progressively. The detailed denoising process is illustrated in the supplementary material. 





\subsection{Training}

In the training stage, the observation and prediction sequences are both available. Thus, we trained the model on the full-motion sequence $X_{full} \in \mathbb{R}^{3 \times (j+1) \times (H+F)}$ . First, we also transferred $X_{full}$ into the DCT space $Y_{full} \in \mathbb{R}^{3 \times (j+1) \times L}$. We then added noise to $Y_{full}$ to generate a noisy sequence $Y_{full}^{t}$. Then we predicted the noise through \autoref{nosieprediciton}.

We then optimised all parameters in our pipeline by minimising the $l_2$ 
loss between the predicted noise $\boldsymbol{\epsilon}_{\boldsymbol{\theta}}\left( Y^t, c, t\right)$ and the true noise $\epsilon$:
\begin{equation}
    \mathcal{L}=\mathbb{E}_{\boldsymbol{\epsilon}, t}\left[\left\|\boldsymbol{\epsilon}-\boldsymbol{\epsilon}_{\boldsymbol{\theta}}\left( Y_{full}^t, c, t\right)\right\|^2\right].
\end{equation}

The detailed training and inference procedure is illustrated in the supplementary material.








\section{Experiments}

\subsection{Datasets}
\label{datasets}
Only a limited number of datasets contain synchronised recordings of both eye gaze and full-body human motion. 
We evaluated our method on two such public datasets, i.e. the MoGaze~\cite{kratzer2020mogaze} dataset for real-world setting and the GIMO~\cite{zheng2022gimo} dataset for AR setting.

\textbf{MoGaze.} The MoGaze dataset provides motion capture and eye-tracking data recorded simultaneously from 6 participants performing $pick$ and $place$ actions in an indoor environment.
It contains over 3 hours of body movement and eye gaze recordings captured at 30 Hz. The pose of the human body is expressed using the 3D coordinates 
of 21 body joints while the eye gaze is represented as a direction vector. 
Following common settings in stochastic human motion prediction~\cite{chen2023humanmac,barquero2022belfusion,yuan2020dlow}, we used recordings from $p1$, $p2$, $p4$, $p5$, and $p6$ for training and employed the data from $p7$ for testing.

\textbf{GIMO.} The GIMO dataset contains body motion and eye gaze data captured from 11 participants in various indoor AR environments.
The action categories include ${sitting}$ or ${laying}$ ${on}$ ${objects}$, ${touching}$, ${holding}$, ${reaching}$ ${to}$ ${objects}$, $opening$, $pushing$, $transferring$, $throwing$, $swapping$ $objects$, etc. 
We represented the body pose in the GIMO dataset using the 3D coordinates of $23$ body joints and denoted the eye gaze using a direction vector.
In the evaluation process, we followed the official dataset splits provided in GIMO~\cite{zheng2022gimo}. 
The training set comprises motion and eye gaze recordings from 12 scenes, while the test set contains data from 14 scenes, including 12 known environments and 2 new, unseen environments. 
This evaluation protocol allows the assessment of generalisation capabilities to new scenes.

\subsection{Evaluation Metrics}
To evaluate the performance of our model, we employed four commonly used metrics following prior works on stochastic HMP~\cite{yuan2020dlow,saadatnejad2023generic,chen2023humanmac}:


\begin{itemize}
    \item Average displacement error (ADE): ADE measures the average $l_2$ distance between the ground truth and the predicted motions over the whole future sequence. The lower the ADE, the more precise the prediction.
    
    \item Final displacement error (FDE): FDE calculates the $l_2$ distance between the ground truth and the predicted future motions at the final timestep.
    The lower the FDE, the more precise the prediction at the final timestep.
    
    \item Multi-modal average displacement error (MMADE): MMADE is designed to handle the multi-modal nature of the predictions in the task of stochastic HMP. It takes into account the fact that there can be multiple reasonable ground truth sequences for a given input. The multiple ground truth is obtained from the future motions of similar observed sequences (Sequences with $l_2$ distance below a given threshold are treated as similar sequences). MMADE is the mean value of the ADE calculated using the predictions and the multiple ground truth.
    The lower the MMADE, the better the ability to generate multiple reasonable predictions.
    
    \item Multi-modal final displacement error (MMFDE): MMFDE is calculated in a similar way as MMADE. The difference is that MMADE calculates the average $l_2$ distance over all future timesteps while MMFDE measures the $l_2$ distance only at the final timestep.
    The lower the MMFDE, the better the ability to generate multiple reasonable predictions at the final timestep.
\end{itemize}



\subsection{Baselines}
We compared our method with the following state-of-the-art methods in stochastic human motion prediction: 
\begin{itemize}    
    \item \textit{DLow}~\cite{yuan2020dlow}: \textit{DLow} is a latent flow-based model to generate multiple future motions via diversity-promoting sampling and loss.
    
    \item \textit{CVAE}~\cite{yuan2020dlow}: \textit{CVAE} is a conditional variational autoencoder utilised in DLow~\cite{yuan2020dlow} as a pre-trained generative model which can also forecast stochastic future motions.
    
    \item \textit{BeLFusion}\cite{barquero2022belfusion}: \textit{BeLFusion} is a latent diffusion-based model to predict reasonable motions based on 
    disentangling the behavioural representation from past motions.
    
    \item \textit{HumanMAC}\cite{chen2023humanmac}: \textit{HumanMac} is a diffusion-based stochastic HMP model with a DCT completion fashion in the inference.
\end{itemize}

\subsection{Implementation Details}

We set the observation and prediction time windows to 0.5 seconds (15 frames) and 2 seconds (60 frames) respectively for both the MoGaze and GIMO datasets, following common practice in stochastic human motion prediction~\cite{chen2023humanmac,barquero2022belfusion,yuan2020dlow}.
For a fair comparison, we retrained all the baseline methods from scratch using their default training parameters. 
We trained all the modules in our method in an end-to-end manner for 300 epochs using Adam optimiser\cite{kingma2014adam} with an initial learning rate of 0.0003 and a batch size of 32.
The learning rate was decayed at 75, 150, 225, 275 epochs, respectively, with a multiplicative factor of 0.9.
We applied a standard diffusion process as proposed by~\cite{song2020denoising} that degraded 1500 steps in the training and sampled 100 steps in the inference. 
We selected the Cosine noise scheduler~\cite{nichol2021improved} in the diffusion 
following HumanMAC~\cite{chen2023humanmac}. 
The multi-modal ground truth 
threshold was set to 0.4 following previous work~\cite{barquero2022belfusion}.
All experiments were conducted in an Nvidia TITAN X GPU with 12GB memory using the PyTorch 1.7.1 framework.

\subsection{Quantitative Results}

\begin{table*}[ht]
  \caption{Comparison of our method with the state-of-the-art methods on both MoGaze~\cite{kratzer2020mogaze} and GIMO~\cite{zheng2022gimo} (unit: meters). The best results are in bold while the second best are underlined.} 
  \label{tab:quantitative_result}
  \scriptsize%
	\centering%
  \begin{tabu}{%
	l%
	*{4}{c}
        |c
        *{3}{c}%
	}
  \toprule
  & \multicolumn{4}{c}{Results on MoGaze~\cite{kratzer2020mogaze}} & \multicolumn{4}{c}{Results on GIMO~\cite{zheng2022gimo}} \\
  \midrule
    &  ADE~$\downarrow$ &   FDE~$\downarrow$ & MMADE~$\downarrow$ & MMFDE~$\downarrow$ &ADE~$\downarrow$ &   FDE~$\downarrow$ & MMADE~$\downarrow$ & MMFDE~$\downarrow$\\
  \midrule
    \textit{CVAE}\cite{yuan2020dlow} &  1.070 &   1.644 & 1.091 & 1.667 &    1.292 &   2.059 & 1.294 & 2.054\\
    
    \textit{DLow}\cite{yuan2020dlow} &  0.807 &   1.261 & 0.840 & 1.274 &  1.084 &   1.688 & 1.090 & 1.687\\
     
     \textit{BeLFusion}\cite{barquero2022belfusion} &  0.899 &   1.306 & 0.908 & 1.313 &  0.840 &   1.220 & 0.845 & 1.221\\
     
       \textit{HumanMAC} \cite{chen2023humanmac} &  0.732 &   1.089 & 0.779 & 1.144  &  0.815 &   1.121 & 0.821 & 1.118\\
      
        \midrule
        \textit{$Ours_{w/o \ gaze}$} &  \underline{0.664} &  0.982 & 0.696 & 0.996&  0.812 & 1.120 & 0.818 & 1.116\\
        \textcolor{black}{\textit{$Ours_{head}$}}&  \textcolor{black}{0.669} &  \textcolor{black}{\underline{0.971}}& \textcolor{black}{\underline{0.678}} & \textcolor{black}{\underline{0.974}} &  \textcolor{black}{\underline{0.734}} & \textcolor{black}{\underline{0.979}} & \textcolor{black}{\underline{0.737}} & \textcolor{black}{\underline{0.975}}\\

         \textit{Ours} &  \textbf{0.638} &  \textbf{ 0.939} & \textbf{0.649} & \textbf{0.946 } &  \textbf{0.729} & \textbf{ 0.974} & \textbf{0.732} & \textbf{0.969}\\
  
  \bottomrule
  \end{tabu}%
\end{table*}

The quantitative results on the MoGaze~\cite{kratzer2020mogaze} and GIMO~\cite{zheng2022gimo} datasets are shown in \autoref{tab:quantitative_result}.
Overall, our model outperforms the state-of-the-art methods on all metrics. 

\textbf{Results on MoGaze.} As shown in \autoref{tab:quantitative_result}, our model achieves the lowest ADE, FDE, MMADE and MMFDE compared to prior methods, substantially surpassing the best state-of-the-art HumanMAC~\cite{chen2023humanmac}. For average displacement error, our method achieves an improvement of 12.8\% (0.638 $vs.$ 0.732) over HumanMAC\cite{chen2023humanmac}. On final displacement error, our method obtains a 13.7\% performance gain (0.939 $vs.$ 1.089). 
On multi-modal accuracy metrics, our method improves the MMADE performance by 16.7\% (0.649 $vs.$ 0.779) and the MMFDE performance by 17.3\% (0.946 $vs.$ 1.144). The greater gains on the multi-modal metrics reveal that our model is more capable of generating multiple reasonable predictions aligned with the stochasticity of human nature.
We further conducted a Wilcoxon signed-rank test to compare our method and the state-of-the-art and the results demonstrate that the differences between our method and the state-of-the-art method are statistically significant ($p<0.01$).
In addition, we also presented the performance of our method without using eye gaze in Table~\ref{tab:quantitative_result}. Compared to our method, the ablated version of not using eye gaze presents lower accuracy in terms of ADE, FDE, MMADE and MMFDE, demonstrating the effectiveness of eye gaze information for stochastic human motion prediction.
It is also notable that our method without using eye gaze still outperforms the state-of-the-art methods, achieving an improvement of 9.8\% on MMADE and 12.9\% 
on MMFDE.
These results underscore the superiority of our model architecture. Considering that eye gaze is not always available, we further present a novel variant that replaces gaze with head direction. We can see from Table~\ref{tab:quantitative_result} that although this variant is not as good as our full model, it outperforms the variant without gaze. These results demonstrate that head direction can be a reliable alternative when gaze is not available.

\textbf{Results on GIMO.} As shown in \autoref{tab:quantitative_result}, our method outperforms the state-of-the-art methods in all the metrics.
Specifically, compared to the state-of-the-art method HumanMAC, our method achieves an improvement of 10.5\% (0.815 to 0.729) in ADE as well as an improvement of 13.1\% (1.121 to 0.974) in FDE.
On multi-modal accuracy metrics, our method obtains an improvement of 10.8\% in MMADE and 13.3\% in MMFDE, respectively.
The results from a Wilcoxon signed-rank test validate that the differences between our method and the state-of-the-art method are statistically significant ($p<0.01$).
We also find that our method significantly outperforms the ablated version of not using eye gaze, revealing the importance of eye gaze for generating stochastic human motions. \textcolor{black}{The results of $Ours_{head}$ on GIMO also validate that eye gaze performs better than head orientation in terms of human motion prediction and revealed that head orientation can be used as a proxy to gaze when gaze is not available.}
\subsection{Visualisation Analysis}
\label{visulisationAn}



We visualised the multiple predicted poses at the time point of the future one second.
We compared our method with 
the state-of-the-art method HumanMAC~\cite{chen2023humanmac}.
For each method, we randomly generated 10 predictions for comparison.

\begin{figure*}[ht]
  \centering
    \includegraphics[width=\textwidth]{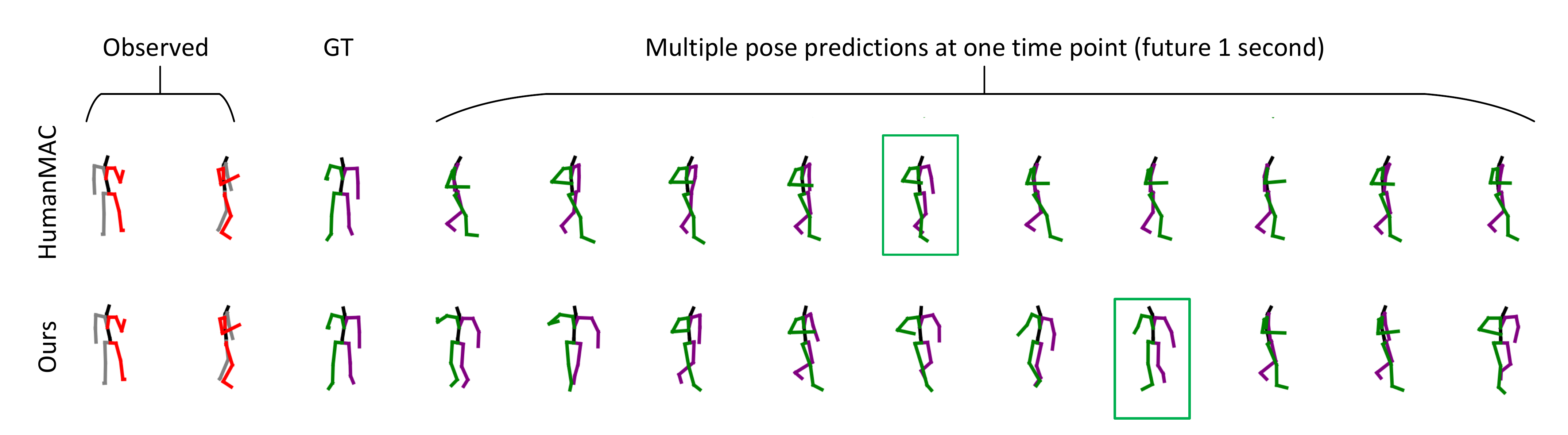}
\caption{Ground truth (GT) human pose at future one second and multiple pose predictions generated by different methods on the MoGaze dataset~\cite{kratzer2020mogaze} with the best prediction (lowest $l_2$ distance to GT) boxed in green.
Our method can generate motions that are closer to the ground truth than the state-of-the-art method HumanMAC~\cite{chen2023humanmac}.}
\label{fig:endpose_precise_case}
\end{figure*}

\textbf{Visualisation results on the MoGaze dataset.} We illustrated a representative visualisation from the MoGaze dataset \cite{zheng2022gimo} in \autoref{fig:endpose_precise_case}. The observed motion sequence involves a turn to the right, and the ground truth pose shows that at the future one second, this person continues the turn by about 100 degrees further to the right. We can observe that predictions of HumanMAC\cite{chen2023humanmac} 
generally continue walking in the same forward direction as the observed last frame, failing to anticipate the full turning trajectory.  The best prediction generated by HumanMAC~\cite{chen2023humanmac} (labelled in green) also differs from the ground truth. In contrast, our method can recognise that the person intends to keep turning right. As a result, the best prediction from our method is precisely facing the true direction of the ground truth and other predictions generally align with the ground truth. 
In addition, we can see that the poses generated by our method are physically plausible without any angle distortions or strange limb lengths.


\textbf{Visualisation results on the GIMO dataset.} We also illustrated a typical visualisation from the GIMO dataset~\cite{zheng2022gimo}. As depicted in \autoref{fig:teaser}, predictions from our method are more similar to the ground truth. Furthermore, all predictions from our method are generally realistic. In contrast, HumanMAC yielded predictions that include some implausible cases (boxed in red). These results demonstrate that our method can generate more reasonable motion predictions than the state-of-the-art.

\begin{figure*}[ht]
  \centering
    \includegraphics[width=\textwidth]{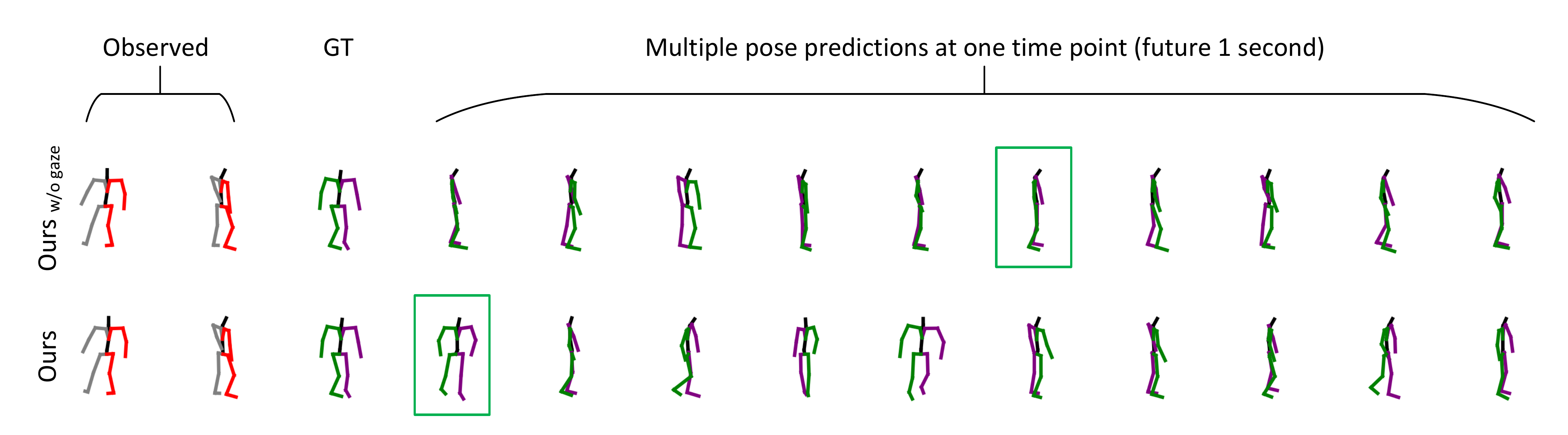}
\caption{Ground truth (GT) human pose at future one second and multiple pose predictions generated by different methods on the MoGaze dataset~\cite{kratzer2020mogaze} with the best prediction (lowest $l_2$ distance to GT) boxed in green. Our method using eye gaze can generate motions that are closer to the ground truth than the ablated version of not using eye gaze.}
\label{fig:endpose_wo_gaze}
\end{figure*}

\textbf{Visualisation results of not using eye gaze.} To further evaluate the effectiveness of eye gaze for stochastic human motion prediction, we showed a visualisation example to compare our method with the ablated version of not using eye gaze.
As shown in \autoref{fig:endpose_wo_gaze}, the observed motions showed a person turning to the right and then standing still. 
Our method was able to predict that the person would continue going right in the future while the ablated version of not using eye gaze failed to recognise this intention and stayed in place.
These results demonstrate that eye gaze signals provide information on human intention and can help improve the performance of stochastic human motion prediction methods.


More visualisation results are provided in the supplementary material.

\subsection{Adding Eye Gaze to Prior Methods}\label{sec:addinggaze}

To further evaluate our method's effectiveness for extracting and fusing gaze and motion features, we compared our methods with prior methods' variants that add eye gaze as input. The results on MoGaze and GIMO are shown in \autoref{tab:quantitative_result_addinggaze}, where $Hum_{gaze}$ and $BeL_{gaze}$ denotes the variants of HumanMAC and BeLFusion that add eye gaze as an extra input joint. Overall, our approach outperforms both BeLFusion and HumanMAC which incorporate eye gaze information in terms of all metrics, demonstrating our method's superiority in exploiting eye gaze information.
Additionally, compared to the performance of original HumanMAC and BeLFusion shown in \autoref{tab:quantitative_result}, all the metrics improve on the MoGaze dataset while getting worse slightly on the GIMO dataset.
It demonstrates that simply adding eye gaze as a new joint to an existing method does not always have a positive effect. 
Instead, the internal features of the eye gaze and its synergistic correlations with other human joint nodes need to be considered. 
Our approach outperforms the state-of-the-art methods significantly by introducing novel feature extraction and fusion modules for gazes and motions and using the fused features to guide future motion generation. 

\begin{table}[ht]
  \caption{Comparison of our method with baselines that add eye gaze as input on both MoGaze~\cite{kratzer2020mogaze} and GIMO~\cite{zheng2022gimo} (unit: meters). The best results are in bold while the second best are underlined.} 
  \label{tab:quantitative_result_addinggaze}
  \scriptsize%
	\centering%
  \begin{tabu}{%
	*{6}{l}%
	}

  \toprule
    Dataset & Method&  ADE &   FDE & MMADE & MMFDE \\
  \midrule
    \multirow{3}{*}{MoGaze}
   &$Hum_{gaze}$ &\underline{0.654} &   \underline{0.965} & 0.679 & \underline{0.980} \\ 
   &$BeL_{gaze}$ &\underline{0.654} &   1.018 & \underline{0.659} & 1.012 \\
   &\textit{Ours} &\textbf{0.638} &  \textbf{ 0.939} & \textbf{0.649} &\textbf{0.946 } \\
   \midrule
   \multirow{3}{*}{GIMO}
   & $Hum_{gaze}$ &\underline{0.850} &   \underline{1.135} & 0.852 & \underline{1.137} \\ 
   &$BeL_{gaze}$ &  \underline{0.858} &   1.126 & \underline{0.861} & 1.221 \\
   &\textit{Ours} &\textbf{0.729} &  \textbf{ 0.974} & \textbf{0.732} &\textbf{0.969 } \\
  \bottomrule
  \end{tabu}%
\end{table}

\subsection{Ablation Study}
We conducted an ablation study to comprehensively analyse the individual contributions of various modules within our framework. 
We thoroughly assessed the performance of our model by comparing it to variants that exclude specific modules:
\begin{itemize}
    \item  $Ours_{w/o \ gat}$ : Without spatio-temporal graph attention network.
    \item $Ours_{w/o \ CA}$ : Without cross attention in the noise predicting network.
    \item $Ours_{w/o \ GE}$ : Without gaze encoder.
    \item $Ours_{w/o \ ME}$ : Without motion encoder.
\end{itemize}


As we can see from \autoref{tab: ablation_study}, our model outperforms all the ablated versions. 
It reveals that each module within our framework significantly contributes to the generation of stochastic human future motions.

%
\begin{table}[H]
  \caption{Ablation study on MoGaze\cite{kratzer2020mogaze}. The best results are in bold.}
  \label{tab: ablation_study}
  \scriptsize%
    \centering%
\resizebox{0.47\textwidth}{!}{
  \begin{tabu}{%
	l%
	*{4}{c}%
	}

  \toprule
    &  ADE~$\downarrow$ &   FDE~$\downarrow$ & MMADE~$\downarrow$ & MMFDE~$\downarrow$ \\
  \midrule

   $Ours_{w/o \ gat}$ &	0.655&	0.990&	0.663&	0.993 \\ 
   $Ours_{w/o \ CA}$ &	0.697&	1.001&	0.709&	1.000  \\
   $Ours_{w/o \ GE}$ &	0.655&	0.946&	0.665&	\textbf{0.946} \\
   $Ours_{w/o \ ME}$ &	0.642&	0.976&	0.654&	0.985 \\
   \midrule
   $Ours$  & \textbf{ 0.638} &  \textbf{0.939} & \textbf{0.649 }& \textbf{0.946} \\
  \bottomrule
  \end{tabu}%
  }
\end{table}

	%



\section{User Study}
To perform a comprehensive evaluation of the perceived precision and realism of the generated samples, we conducted a user study to assess model performance based on human perception and intuition. Participants were tasked with random samples predicted by different motion forecasting models. By collecting subjective rankings directly from individuals, the user study provided an intuitive approach to evaluate the aspects of the predictions that are challenging to assess objectively, such as naturalness, continuity, and the overall plausibility of the motions.

\subsection{Assessment Details}
We randomly selected 4 sequences from the MoGaze dataset and 12 sequences from the GIMO dataset.
The difference in the number of selections between the two datasets can be attributed to the difference in the variety of actions included. While the MoGaze dataset consists of only two actions, the GIMO dataset contains a larger number of different actions. We compared our method with the best baseline HumanMAC. \textcolor{black}{The compared conditions were randomly provided to the users in order to avoid any potential bias.} Each method generated 10 predictions from random noise. \textcolor{black}{We set the sample size to 10 for a trade-off between the predictions’ diversity and users’ burden}. The participants in our study were then instructed to rank these predictions based on the following two key aspects:

\begin{itemize}
\item \textbf{Realism}: If these poses are physically plausible. You can check if there are any angle distortions, too short/long limbs, implausible poses, or any sudden or unreasonable changes during the whole motion.

\item \textbf{Precision}: If these motions align with the ground truth. You can measure the similarity between each motion and the ground truth.
\end{itemize}

The questionnaire utilised in our study was created using the \textit{Jotform} platform
The interface of our user study can be found in the supplementary material. A total of 21 individuals (11 males and 10 females, aged between 21
and 39 years, Mean= 26.3, SD=4.4) were recruited to take part in our user study via university mailing lists and social networks.
This user study is approved by the university.

\subsection{Evaluation}

As shown in \autoref{tab: user study}, our method outperforms HumanMAC in terms of both precision and realism. 74.6\% of the total responses considered predictions of our method as more precise while only 25.4\% regarded HumanMAC's results as more precise. From the perspective of realism, 67.7\% of the total responses considered the motions generated by our method as more realistic than HumanMAC
%
while only 33.3\% of the responses considered HumanMAC were able to generate more realistic predictions. 
To further validate our findings, we conducted a paired Wilcoxon signed-rank test, which revealed a statistically significant difference ($p<0.01$) between our method and HumanMAC in terms of both precision and realism.

\begin{table}[H]
  \caption{Results of the user study. Participants were required to measure the predictions from different methods according to precision and realism.}
  \label{tab: user study}
  \scriptsize%

	\centering%
  \begin{tabu}{%
	l%
	*{4}{c}%
	}
  \toprule
    & More precise & More realistic   \\
  \midrule
   $HumanMAC$ & 25.4\% & 32.3\%\\
   $Ours$ &74.6\%&	67.7\%  \\

  \bottomrule
  \end{tabu}%
\end{table}
\section{Discussion}
Our study marks a notable advancement in the emerging field of incorporating eye gaze cues for stochastic human motion prediction in VR/AR research. We have successfully demonstrated the effectiveness of our proposed method in generating multiple reasonable human motion predictions.


\textbf{Performance of our method.} Our method outperforms the state-of-the-art methods in terms of all metrics on both the MoGaze and GIMO datasets (See \autoref{tab:quantitative_result}). In addition, the visualisation results on both datasets showed that our method can generate multiple reasonable predictions with high precision while the state-of-the-art may produce erroneous or implausible predictions (\autoref{fig:teaser} and \autoref{fig:endpose_precise_case}).
The results from a user study further confirmed that the predictions generated by our method were more realistic and more precise than that from prior method (\autoref{tab: user study}). These results demonstrate that our work have made a significant improvement in stochastic human motion prediction in terms of improving both precision and realism.

\textbf{The effectiveness of eye gaze for stochastic HMP.} To the best of our knowledge, our method is the first to introduce eye gaze in the stochastic HMP task to obtain a significant performance improvement.
Compared to the full model, the ablated version of not using eye gaze performed significantly worse on both datasets (\autoref{tab:quantitative_result}). It demonstrated that introducing the historical eye gaze as input can benefit the generation of more precise future motions. The visualisation example further demonstrated the effectiveness of incorporating eye gaze.
As shown in \autoref{fig:endpose_wo_gaze}, the observed motion showed a person turned to the right and then standing. The ground truth pose at future one second showed the person turning right again from a standstill. Our full model with eye gaze can predict this intention of turning right while the ablated version of not using eye gaze failed to recognise it and generated predictions that were all staying in place. 
We argue that the eye gaze implies the intention of future body movements. Therefore, our method can recognise more complex and precise intentions and make reasonable and precise predictions.

\textbf{The approach to introducing gaze into stochastic HMP.} Our method showed a significant improvement in the stochastic HMP task by adding eye gaze as input. A natural question is whether adding eye gaze as additional input to an existing method has an effect. In \autoref{sec:addinggaze} we explored existing diffusion-based methods with eye gaze as an additional joint input. The accuracy declined slightly on the GIMO dataset (\autoref{tab:quantitative_result_addinggaze}). Compared to the MoGaze dataset, the GIMO dataset contains more complex actions and motions, and the correlation between eye gaze and future action intent is much harder to mine. Therefore, simply adding eye gaze as an additional joint makes it difficult to model this complex correlation. Our method extracted the gaze and motion features via two encoders and fused them via GAT. Experimental results demonstrated the effectiveness of our gaze encoder, motion encoder, as well as the GAT (\autoref{tab: ablation_study}).


\textcolor{black}{\textbf{Diversity reduction due to eye gaze guidance.} As shown in \autoref{tab:quantitative_result}, our model reported obvious improvements in terms of all accuracy metrics. Intuitively, the diversity of predictions will decrease. To further explore whether and to what extent the diversity of our predictions were reduced, we used the average pairwise distance (APD), i.e. the average L2 distance between every pair of motion predictions, to measure the diversity. We observed that when ablating eye gaze from our model, the APD value increases from 15.572 to 15.799 on MoGaze and from 15.519 to 20.199 on GIMO. These results validate that while adding eye gaze can improve accuracy, it inevitably decreases the diversity of the motion predictions. 
}

\textbf{Limitations.} 
First, only two public datasets contain both the eye gaze and body motions, limiting our evaluation's generalisability. In addition, existing datasets contain limited action categories, which make predictions that tend to be limited to a few number of actions and can not generalise to unseen actions.
. Finally, our approach uses observed motions and eye gaze as input. However, the input data may be missing due to tracking errors in practical applications, which may limit the application of our method in real-world scenarios.


\textbf{Future work.}  Our proposed approach, GazeMoDiff, has the potential to be extended to incorporate additional modalities beyond eye gaze. It serves as a multi-modal human motion prediction framework, where various guidance information can be utilised to enhance the generation of reasonable motions. By incorporating modalities such as facial expressions, hand gestures, and physiological signals, the GazeMoDiff framework can be expanded to capture a more comprehensive range of cues for motion prediction. In addition, we are also looking forward to integrating our method with real VR/AR applications. Finally, as a diffusion-based prediction model, our method needs 100 DDIM sampling steps to denoise and obtain predictions, which limits to implementation into real-time scenarios. We will try some fast sampling methods~\cite{lu2022dpm,lu2022dpm++} to improve generation speed. 
\section{Conclusion}

In this work, we were the first to explore the effectiveness of eye gaze on generating multiple reasonable human motions in the future.
We proposed a novel gaze-guided diffusion model that fuses the gaze and motion features using a spatio-temporal graph attention network and then injects these features into a noise prediction network via a cross-attention mechanism to generate multiple reasonable human future motions. 
Extensive experiments demonstrated that our method outperforms the state-of-the-art methods by a large margin 
and a user study validated that our method can generate motions that are more precise and more realistic than prior methods.
As such, our work makes an important step towards generating more realistic human motions for virtual agents and guides future work on cross-modal human behaviour generation.

\section*{Acknowledgments}
This work was funded by the Deutsche Forschungsgemeinschaft (DFG, German Research Foundation) under Germany's Excellence Strategy -- EXC 2075 -- 390740016.

\bibliographystyle{abbrv-doi}

\bibliography{template}

\begin{thebibliography}{10}

\bibitem{aksan2021spatio}
E.~Aksan, M.~Kaufmann, P.~Cao, and O.~Hilliges.
\newblock A spatio-temporal transformer for 3d human motion prediction.
\newblock In {\em 2021 International Conference on 3D Vision (3DV)}, pp.
  565--574. IEEE, 2021.

\bibitem{aliakbarian2021contextually}
S.~Aliakbarian, F.~Saleh, L.~Petersson, S.~Gould, and M.~Salzmann.
\newblock Contextually plausible and diverse 3d human motion prediction.
\newblock In {\em Proceedings of the IEEE/CVF International Conference on
  Computer Vision}, pp. 11333--11342, 2021.

\bibitem{azmandian2016automated}
M.~Azmandian, T.~Grechkin, M.~Bolas, and E.~Suma.
\newblock Automated path prediction for redirected walking using navigation
  meshes.
\newblock In {\em 2016 IEEE Symposium on 3D User Interfaces (3DUI)}, pp.
  63--66. IEEE, 2016.

\bibitem{barquero2022belfusion}
G.~Barquero, S.~Escalera, and C.~Palmero.
\newblock Belfusion: Latent diffusion for behavior-driven human motion
  prediction.
\newblock {\em arXiv preprint arXiv:2211.14304}, 2022.

\bibitem{barsoum2018hp}
E.~Barsoum, J.~Kender, and Z.~Liu.
\newblock Hp-gan: Probabilistic 3d human motion prediction via gan.
\newblock In {\em Proceedings of the IEEE conference on computer vision and
  pattern recognition workshops}, pp. 1418--1427, 2018.

\bibitem{bhattacharya2021text2gestures}
U.~Bhattacharya, N.~Rewkowski, A.~Banerjee, P.~Guhan, A.~Bera, and D.~Manocha.
\newblock Text2gestures: A transformer-based network for generating emotive
  body gestures for virtual agents.
\newblock In {\em 2021 IEEE virtual reality and 3D user interfaces (VR)}, pp.
  1--10. IEEE, 2021.

\bibitem{butepage2017deep}
J.~Butepage, M.~J. Black, D.~Kragic, and H.~Kjellstrom.
\newblock Deep representation learning for human motion prediction and
  classification.
\newblock In {\em Proceedings of the IEEE conference on computer vision and
  pattern recognition}, pp. 6158--6166, 2017.

\bibitem{cao2020long}
Z.~Cao, H.~Gao, K.~Mangalam, Q.-Z. Cai, M.~Vo, and J.~Malik.
\newblock Long-term human motion prediction with scene context.
\newblock In {\em Computer Vision--ECCV 2020: 16th European Conference,
  Glasgow, UK, August 23--28, 2020, Proceedings, Part I 16}, pp. 387--404.
  Springer, 2020.

\bibitem{9983833}
H.~Chen, L.~Wei, H.~Liu, B.~Shi, G.~Zhang, and H.~Zha.
\newblock Mount: Learning 6dof motion prediction based on uncertainty
  estimation for delayed ar rendering.
\newblock {\em IEEE Transactions on Visualization and Computer Graphics}, pp.
  1--12, 2022. doi: {{%
10\hspace{.1pt}\discretionary{.}{%
}{.}\hspace{.4pt}1109\discretionary{/}{%
}{/}TVCG\hspace{.1pt}\discretionary{.}{%
}{.}\hspace{.4pt}2022\hspace{.1pt}\discretionary{.}{%
}{.}\hspace{.4pt}3228807}}


\bibitem{chen2023humanmac}
L.-H. Chen, J.~Zhang, Y.~Li, Y.~Pang, X.~Xia, and T.~Liu.
\newblock Humanmac: Masked motion completion for human motion prediction.
\newblock {\em arXiv preprint arXiv:2302.03665}, 2023.

\bibitem{chen2022diffusiondet}
S.~Chen, P.~Sun, Y.~Song, and P.~Luo.
\newblock Diffusiondet: Diffusion model for object detection.
\newblock {\em arXiv preprint arXiv:2211.09788}, 2022.

\bibitem{david2021towards}
B.~David-John, C.~Peacock, T.~Zhang, T.~S. Murdison, H.~Benko, and T.~R.
  Jonker.
\newblock Towards gaze-based prediction of the intent to interact in virtual
  reality.
\newblock In {\em ACM Symposium on Eye Tracking Research and Applications}, pp.
  1--7, 2021.

\bibitem{dhariwal2021diffusion}
P.~Dhariwal and A.~Nichol.
\newblock Diffusion models beat gans on image synthesis.
\newblock {\em Advances in Neural Information Processing Systems},
  34:8780--8794, 2021.

\bibitem{du2023avatars}
Y.~Du, R.~Kips, A.~Pumarola, S.~Starke, A.~Thabet, and A.~Sanakoyeu.
\newblock Avatars grow legs: Generating smooth human motion from sparse
  tracking inputs with diffusion model.
\newblock In {\em Proceedings of the IEEE/CVF Conference on Computer Vision and
  Pattern Recognition}, pp. 481--490, 2023.

\bibitem{emery2021openneeds}
K.~J. Emery, M.~Zannoli, J.~Warren, L.~Xiao, and S.~S. Talathi.
\newblock Openneeds: A dataset of gaze, head, hand, and scene signals during
  exploration in open-ended vr environments.
\newblock In {\em ACM Symposium on Eye Tracking Research and Applications}, pp.
  1--7, 2021.

\bibitem{fragkiadaki2015recurrent}
K.~Fragkiadaki, S.~Levine, P.~Felsen, and J.~Malik.
\newblock Recurrent network models for human dynamics.
\newblock In {\em Proceedings of the IEEE international conference on computer
  vision}, pp. 4346--4354, 2015.

\bibitem{freedman2008coordination}
E.~G. Freedman.
\newblock Coordination of the eyes and head during visual orienting.
\newblock {\em Experimental brain research}, 190:369--387, 2008.

\bibitem{gamage2021so}
N.~M. Gamage, D.~Ishtaweera, M.~Weigel, and A.~Withana.
\newblock So predictable! continuous 3d hand trajectory prediction in virtual
  reality.
\newblock In {\em The 34th Annual ACM Symposium on User Interface Software and
  Technology}, pp. 332--343, 2021.

\bibitem{ghosh2017learning}
P.~Ghosh, J.~Song, E.~Aksan, and O.~Hilliges.
\newblock Learning human motion models for long-term predictions.
\newblock In {\em 2017 International Conference on 3D Vision (3DV)}, pp.
  458--466. IEEE, 2017.

\bibitem{gopalakrishnan2019neural}
A.~Gopalakrishnan, A.~Mali, D.~Kifer, L.~Giles, and A.~G. Ororbia.
\newblock A neural temporal model for human motion prediction.
\newblock In {\em Proceedings of the IEEE/CVF Conference on Computer Vision and
  Pattern Recognition}, pp. 12116--12125, 2019.

\bibitem{ho2020denoising}
J.~Ho, A.~Jain, and P.~Abbeel.
\newblock Denoising diffusion probabilistic models.
\newblock {\em Advances in neural information processing systems},
  33:6840--6851, 2020.

\bibitem{ho2022classifier}
J.~Ho and T.~Salimans.
\newblock Classifier-free diffusion guidance.
\newblock {\em arXiv preprint arXiv:2207.12598}, 2022.

\bibitem{ho2022video}
J.~Ho, T.~Salimans, A.~Gritsenko, W.~Chan, M.~Norouzi, and D.~J. Fleet.
\newblock Video diffusion models.
\newblock {\em arXiv preprint arXiv:2204.03458}, 2022.

\bibitem{holm2012collision}
J.~E. Holm.
\newblock {\em Collision prediction and prevention in a simultaneous multi-user
  immersive virtual environment}.
\newblock PhD thesis, Miami University, 2012.

\bibitem{hou2019head}
X.~Hou, J.~Zhang, M.~Budagavi, and S.~Dey.
\newblock Head and body motion prediction to enable mobile vr experiences with
  low latency.
\newblock In {\em 2019 IEEE Global Communications Conference (GLOBECOM)}, pp.
  1--7. IEEE, 2019.

\bibitem{hu2021ehtask}
Z.~Hu, A.~Bulling, S.~Li, and G.~Wang.
\newblock Ehtask: Recognizing user tasks from eye and head movements in
  immersive virtual reality.
\newblock {\em IEEE Transactions on Visualization and Computer Graphics}, 2021.

\bibitem{hu2021fixationnet}
Z.~Hu, A.~Bulling, S.~Li, and G.~Wang.
\newblock Fixationnet: Forecasting eye fixations in task-oriented virtual
  environments.
\newblock {\em IEEE Transactions on Visualization and Computer Graphics},
  27(5):2681--2690, 2021.

\bibitem{hu2020dgaze}
Z.~Hu, S.~Li, C.~Zhang, K.~Yi, G.~Wang, and D.~Manocha.
\newblock Dgaze: Cnn-based gaze prediction in dynamic scenes.
\newblock {\em IEEE Transactions on Visualization and Computer Graphics},
  26(5):1902--1911, 2020.

\bibitem{hu24pose2gaze}
Z.~Hu, J.~Xu, S.~Schmitt, and A.~Bulling.
\newblock Pose2gaze: Eye-body coordination during daily activities for gaze
  prediction from full-body poses.
\newblock {\em IEEE Transactions on Visualization and Computer Graphics}, 2024.

\bibitem{hu24hoimotion}
Z.~Hu, Z.~Yin, D.~Haeufle, S.~Schmitt, and A.~Bulling.
\newblock Hoimotion: Forecasting human motion during human-object interactions
  using egocentric 3d object bounding boxes.
\newblock {\em IEEE Transactions on Visualization and Computer Graphics}, 2024.

\bibitem{hu2019sgaze}
Z.~Hu, C.~Zhang, S.~Li, G.~Wang, and D.~Manocha.
\newblock Sgaze: A data-driven eye-head coordination model for realtime gaze
  prediction.
\newblock {\em IEEE transactions on visualization and computer graphics},
  25(5):2002--2010, 2019.

\bibitem{jain2020gan}
D.~K. Jain, M.~Zareapoor, R.~Jain, A.~Kathuria, and S.~Bachhety.
\newblock Gan-poser: an improvised bidirectional gan model for human motion
  prediction.
\newblock {\em Neural Computing and Applications}, 32(18):14579--14591, 2020.

\bibitem{kim2017activity}
Y.~Kim, J.~An, M.~Lee, and Y.~Lee.
\newblock An activity-embedding approach for next-activity prediction in a
  multi-user smart space.
\newblock In {\em 2017 IEEE International Conference on Smart Computing
  (SMARTCOMP)}, pp. 1--6. IEEE, 2017.

\bibitem{kingma2014adam}
D.~P. Kingma and J.~Ba.
\newblock Adam: A method for stochastic optimization.
\newblock {\em arXiv preprint arXiv:1412.6980}, 2014.

\bibitem{koochaki2018predicting}
F.~Koochaki and L.~Najafizadeh.
\newblock Predicting intention through eye gaze patterns.
\newblock In {\em 2018 IEEE Biomedical Circuits and Systems Conference
  (BioCAS)}, pp. 1--4. IEEE, 2018.

\bibitem{kothari2020gaze}
R.~Kothari, Z.~Yang, C.~Kanan, R.~Bailey, J.~B. Pelz, and G.~J. Diaz.
\newblock Gaze-in-wild: A dataset for studying eye and head coordination in
  everyday activities.
\newblock {\em Scientific reports}, 10(1):2539, 2020.

\bibitem{kratzer2020mogaze}
P.~Kratzer, S.~Bihlmaier, N.~B. Midlagajni, R.~Prakash, M.~Toussaint, and
  J.~Mainprice.
\newblock Mogaze: A dataset of full-body motions that includes workspace
  geometry and eye-gaze.
\newblock {\em IEEE Robotics and Automation Letters}, 6(2):367--373, 2020.

\bibitem{kundu2019bihmp}
J.~N. Kundu, M.~Gor, and R.~V. Babu.
\newblock Bihmp-gan: Bidirectional 3d human motion prediction gan.
\newblock In {\em Proceedings of the AAAI conference on artificial
  intelligence}, vol.~33, pp. 8553--8560, 2019.

\bibitem{li2021multiscale}
M.~Li, S.~Chen, Y.~Zhao, Y.~Zhang, Y.~Wang, and Q.~Tian.
\newblock Multiscale spatio-temporal graph neural networks for 3d
  skeleton-based motion prediction.
\newblock {\em IEEE Transactions on Image Processing}, 30:7760--7775, 2021.

\bibitem{li2021directed}
Q.~Li, G.~Chalvatzaki, J.~Peters, and Y.~Wang.
\newblock Directed acyclic graph neural network for human motion prediction.
\newblock In {\em 2021 IEEE International Conference on Robotics and Automation
  (ICRA)}, pp. 3197--3204. IEEE, 2021.

\bibitem{li2017auto}
Z.~Li, Y.~Zhou, S.~Xiao, C.~He, Z.~Huang, and H.~Li.
\newblock Auto-conditioned recurrent networks for extended complex human motion
  synthesis.
\newblock {\em arXiv preprint arXiv:1707.05363}, 2017.

\bibitem{lu2022dpm}
C.~Lu, Y.~Zhou, F.~Bao, J.~Chen, C.~Li, and J.~Zhu.
\newblock Dpm-solver: A fast ode solver for diffusion probabilistic model
  sampling in around 10 steps.
\newblock {\em Advances in Neural Information Processing Systems},
  35:5775--5787, 2022.

\bibitem{lu2022dpm++}
C.~Lu, Y.~Zhou, F.~Bao, J.~Chen, C.~Li, and J.~Zhu.
\newblock Dpm-solver++: Fast solver for guided sampling of diffusion
  probabilistic models.
\newblock {\em arXiv preprint arXiv:2211.01095}, 2022.

\bibitem{ma2022progressively}
T.~Ma, Y.~Nie, C.~Long, Q.~Zhang, and G.~Li.
\newblock Progressively generating better initial guesses towards next stages
  for high-quality human motion prediction.
\newblock In {\em Proceedings of the IEEE/CVF Conference on Computer Vision and
  Pattern Recognition}, pp. 6437--6446, 2022.

\bibitem{mao2021generating}
W.~Mao, M.~Liu, and M.~Salzmann.
\newblock Generating smooth pose sequences for diverse human motion prediction.
\newblock In {\em Proceedings of the IEEE/CVF International Conference on
  Computer Vision}, pp. 13309--13318, 2021.

\bibitem{mao2019learning}
W.~Mao, M.~Liu, M.~Salzmann, and H.~Li.
\newblock Learning trajectory dependencies for human motion prediction.
\newblock In {\em Proceedings of the IEEE/CVF International Conference on
  Computer Vision}, pp. 9489--9497, 2019.

\bibitem{martinez2017human}
J.~Martinez, M.~J. Black, and J.~Romero.
\newblock On human motion prediction using recurrent neural networks.
\newblock In {\em Proceedings of the IEEE conference on computer vision and
  pattern recognition}, pp. 2891--2900, 2017.

\bibitem{martinez2021pose}
A.~Mart{\'\i}nez-Gonz{\'a}lez, M.~Villamizar, and J.-M. Odobez.
\newblock Pose transformers (potr): Human motion prediction with
  non-autoregressive transformers.
\newblock In {\em Proceedings of the IEEE/CVF International Conference on
  Computer Vision}, pp. 2276--2284, 2021.

\bibitem{nichol2021glide}
A.~Nichol, P.~Dhariwal, A.~Ramesh, P.~Shyam, P.~Mishkin, B.~McGrew,
  I.~Sutskever, and M.~Chen.
\newblock Glide: Towards photorealistic image generation and editing with
  text-guided diffusion models.
\newblock {\em arXiv preprint arXiv:2112.10741}, 2021.

\bibitem{nichol2021improved}
A.~Q. Nichol and P.~Dhariwal.
\newblock Improved denoising diffusion probabilistic models.
\newblock In {\em International Conference on Machine Learning}, pp.
  8162--8171. PMLR, 2021.

\bibitem{ramesh2022hierarchical}
A.~Ramesh, P.~Dhariwal, A.~Nichol, C.~Chu, and M.~Chen.
\newblock Hierarchical text-conditional image generation with clip latents.
\newblock {\em arXiv preprint arXiv:2204.06125}, 1(2):3, 2022.

\bibitem{rasul2021autoregressive}
K.~Rasul, C.~Seward, I.~Schuster, and R.~Vollgraf.
\newblock Autoregressive denoising diffusion models for multivariate
  probabilistic time series forecasting.
\newblock In {\em International Conference on Machine Learning}, pp.
  8857--8868. PMLR, 2021.

\bibitem{rombach2022high}
R.~Rombach, A.~Blattmann, D.~Lorenz, P.~Esser, and B.~Ommer.
\newblock High-resolution image synthesis with latent diffusion models.
\newblock In {\em Proceedings of the IEEE/CVF Conference on Computer Vision and
  Pattern Recognition}, pp. 10684--10695, 2022.

\bibitem{saadatnejad2023generic}
S.~Saadatnejad, A.~Rasekh, M.~Mofayezi, Y.~Medghalchi, S.~Rajabzadeh,
  T.~Mordan, and A.~Alahi.
\newblock A generic diffusion-based approach for 3d human pose prediction in
  the wild.
\newblock In {\em 2023 IEEE International Conference on Robotics and Automation
  (ICRA)}, pp. 8246--8253. IEEE, 2023.

\bibitem{sidenmark2019eye}
L.~Sidenmark and H.~Gellersen.
\newblock Eye, head and torso coordination during gaze shifts in virtual
  reality.
\newblock {\em ACM Transactions on Computer-Human Interaction (TOCHI)},
  27(1):1--40, 2019.

\bibitem{sidenmark2019selection}
L.~Sidenmark and H.~Gellersen.
\newblock Eye\&head: Synergetic eye and head movement for gaze pointing and
  selection.
\newblock In {\em Proceedings of the 2019 ACM Symposium on User Interface
  Software and Technology}, pp. 1161--1174, 2019.

\bibitem{sidenmark2020bimodalgaze}
L.~Sidenmark, D.~Mardanbegi, A.~R. Gomez, C.~Clarke, and H.~Gellersen.
\newblock Bimodalgaze: Seamlessly refined pointing with gaze and filtered
  gestural head movement.
\newblock In {\em ACM Symposium on Eye Tracking Research and Applications}, pp.
  1--9, 2020.

\bibitem{song2020denoising}
J.~Song, C.~Meng, and S.~Ermon.
\newblock Denoising diffusion implicit models.
\newblock {\em arXiv preprint arXiv:2010.02502}, 2020.

\bibitem{sun2018towards}
Q.~Sun, A.~Patney, L.-Y. Wei, O.~Shapira, J.~Lu, P.~Asente, S.~Zhu, M.~McGuire,
  D.~Luebke, and A.~Kaufman.
\newblock Towards virtual reality infinite walking: dynamic saccadic
  redirection.
\newblock {\em ACM Transactions on Graphics (TOG)}, 37(4):1--13, 2018.

\bibitem{tashiro2021csdi}
Y.~Tashiro, J.~Song, Y.~Song, and S.~Ermon.
\newblock Csdi: Conditional score-based diffusion models for probabilistic time
  series imputation.
\newblock {\em Advances in Neural Information Processing Systems},
  34:24804--24816, 2021.

\bibitem{tevet2022human}
G.~Tevet, S.~Raab, B.~Gordon, Y.~Shafir, D.~Cohen-Or, and A.~H. Bermano.
\newblock Human motion diffusion model.
\newblock {\em arXiv preprint arXiv:2209.14916}, 2022.

\bibitem{vaswani2017attention}
A.~Vaswani, N.~Shazeer, N.~Parmar, J.~Uszkoreit, L.~Jones, A.~N. Gomez,
  {\L}.~Kaiser, and I.~Polosukhin.
\newblock Attention is all you need.
\newblock {\em Advances in neural information processing systems}, 30, 2017.

\bibitem{wyatt2022anoddpm}
J.~Wyatt, A.~Leach, S.~M. Schmon, and C.~G. Willcocks.
\newblock Anoddpm: Anomaly detection with denoising diffusion probabilistic
  models using simplex noise.
\newblock In {\em Proceedings of the IEEE/CVF Conference on Computer Vision and
  Pattern Recognition}, pp. 650--656, 2022.

\bibitem{xu2022dream3d}
J.~Xu, X.~Wang, W.~Cheng, Y.-P. Cao, Y.~Shan, X.~Qie, and S.~Gao.
\newblock Dream3d: Zero-shot text-to-3d synthesis using 3d shape prior and
  text-to-image diffusion models.
\newblock {\em arXiv preprint arXiv:2212.14704}, 2022.

\bibitem{yuan2020dlow}
Y.~Yuan and K.~Kitani.
\newblock Dlow: Diversifying latent flows for diverse human motion prediction.
\newblock In {\em Computer Vision--ECCV 2020: 16th European Conference,
  Glasgow, UK, August 23--28, 2020, Proceedings, Part IX 16}, pp. 346--364.
  Springer, 2020.

\bibitem{zangemeister1982gaze}
W.~H. Zangemeister and L.~Stark.
\newblock Gaze latency: variable interactions of head and eye latency.
\newblock {\em Experimental Neurology}, 75(2):389--406, 1982.

\bibitem{zhang2023adding}
L.~Zhang and M.~Agrawala.
\newblock Adding conditional control to text-to-image diffusion models.
\newblock {\em arXiv preprint arXiv:2302.05543}, 2023.

\bibitem{zhang2022motiondiffuse}
M.~Zhang, Z.~Cai, L.~Pan, F.~Hong, X.~Guo, L.~Yang, and Z.~Liu.
\newblock Motiondiffuse: Text-driven human motion generation with diffusion
  model.
\newblock {\em arXiv preprint arXiv:2208.15001}, 2022.

\bibitem{zheng2022gimo}
Y.~Zheng, Y.~Yang, K.~Mo, J.~Li, T.~Yu, Y.~Liu, C.~K. Liu, and L.~J. Guibas.
\newblock Gimo: Gaze-informed human motion prediction in context.
\newblock In {\em European Conference on Computer Vision}, pp. 676--694.
  Springer, 2022.

\end{thebibliography}
\end{document}